\documentclass[10pt,twocolumn,letterpaper]{article}
\usepackage[pagenumbers]{cvpr}
\usepackage{microtype}
\usepackage{xurl}
\usepackage{multirow}
\usepackage{colortbl}
\usepackage{overpic}
\usepackage{needspace}
\definecolor{efficiencygreen}{HTML}{009A55}
\definecolor{tableseparator}{gray}{0.72}
\newcommand{\effvalue}[1]{\textcolor{efficiencygreen}{\textbf{#1}}}
\newcommand{\methodrule}{\arrayrulecolor{tableseparator}\midrule[0.3pt]\arrayrulecolor{black}}
\definecolor{cvprblue}{rgb}{0.21,0.49,0.74}
\usepackage[breaklinks,colorlinks,allcolors=cvprblue]{hyperref}
\newcommand{\proj}{\textsc{CueSplat}\xspace}

\newcommand{\Sect}[1]{Sec.~\ref{#1}}
\newcommand{\Fig}[1]{Fig.~\ref{#1}}
\newcommand{\Tbl}[1]{Tbl.~\ref{#1}}

\hypersetup{
  pdftitle={CueSplat: Learning to Schedule Sparse Deformation for Dynamic Gaussian Splatting on Edge Devices},
  pdfauthor={Qingyuan Tang, He Zhu, Shixuan Sun, Anbang Wu, Jingwen Leng, Minyi Guo, Yu Feng}
}
\graphicspath{{figs/}}
\title{\proj: Learning to Schedule Sparse Deformation for Dynamic Gaussian Splatting on Edge Devices}
\author{%
Qingyuan Tang\textsuperscript{1}\quad
He Zhu\textsuperscript{1}\quad
Shixuan Sun\textsuperscript{1}\quad
Anbang Wu\textsuperscript{1}\\[2pt]
Jingwen Leng\textsuperscript{1}\quad
Minyi Guo\textsuperscript{1}\quad
Yu Feng\textsuperscript{1}\\[4pt]
\textsuperscript{1}Shanghai Jiaotong University\\[3pt]
{\normalsize Project Page: \url{https://cuesplat.github.io/}}
}

\makeatletter
\patchcmd{\@maketitle}{\vskip .375in}{\vskip .125in}{}{\PackageError{paper}{Could not adjust title top spacing}{}}
\patchcmd{\@maketitle}{\vspace*{12pt}}{\vspace*{0pt}}{}{\PackageError{paper}{Could not adjust title spacing}{}}
\patchcmd{\@maketitle}{\vspace*{24pt}}{\vspace*{12pt}}{}{\PackageError{paper}{Could not adjust author spacing}{}}
\apptocmd{\@maketitle}{\begin{minipage}{\textwidth}
\centering
\captionsetup{type=figure,hypcap=false}
\begingroup
\normalfont\fontsize{8}{9.6}\selectfont
\includegraphics[width=\linewidth]{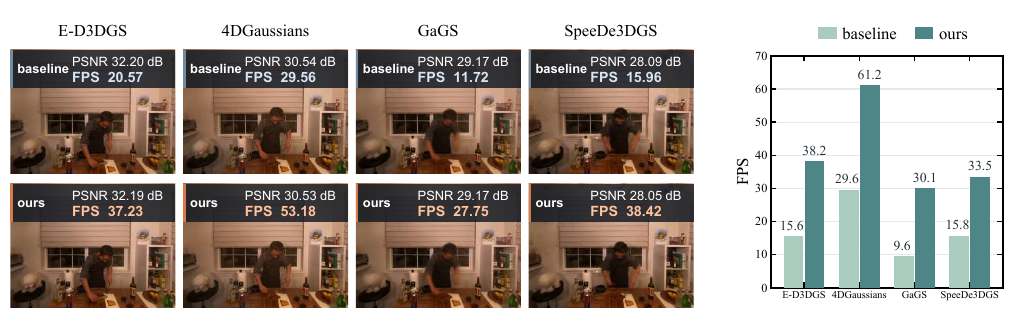}\par
\vspace{3pt}
\noindent\makebox[.68\linewidth][c]{(a) cut\_roasted\_beef: scene-average PSNR (dB) and FPS}%
\makebox[.32\linewidth][c]{(b) Average over 9 N3DV scenes}\par
\endgroup

\caption{\proj{} accelerates four deformation-based Gaussian backbones: E-D3DGS~\cite{bae2024ed3dgs}, 4DGaussians~\cite{wu2024four}, GaGS~\cite{lu2024gags}, and SpeeDe3DGS~\cite{tu2026speede}. (a) Renderings and scene-average PSNR/FPS on \textit{cut\_roasted\_beef}. (b) Average FPS over nine N3DV scenes~\cite{li2022neuralvideo}. Throughput uses the same Jetson AGX Orin measurements as \Tbl{tab:main_results}.}
\label{fig:teaser}
\end{minipage}
}{}{\PackageError{paper}{Could not place the teaser}{}}
\makeatother

\begin{document}
\raggedbottom
\maketitle
\enlargethispage{\baselineskip}
\begingroup
\setlength{\parskip}{0pt}
\begin{abstract}
	Deformation-based 4D Gaussian splatting (4DGS) enables dynamic scene rendering.
	However, evaluating deformation networks for all Gaussians at every frame is compute-intensive on edge devices.
	We find that Gaussian trajectories are often locally predictable, yet prediction reliability varies across Gaussians and time.
	Thus, we propose \proj{}, a post-training framework that exploits local temporal regularity while keeping the pretrained backbone frozen and retaining its Gaussian set.
	Specifically, \proj{} predicts Gaussian attributes from recent history and uses a lightweight learned router to select exact deformation updates within a scene-specific budget calibrated offline.
	Selection accounts for prediction reliability and rendering contribution.
	Selected indices are reused across nearby frames to reduce routing overhead, while selected Gaussians still receive exact updates each frame.
	To control accumulated prediction error, we use drift signals to schedule three consecutive full-deformation frames that rebuild the prediction history.
	Evaluated across four mainstream 4DGS algorithms, \proj achieves up to $6.74\times$ and $8.15\times$ speedups on Jetson AGX Orin and Orin NX, respectively, with a negligible quality drop
	(at most $0.16$~dB in dataset-average PSNR).
	
\end{abstract}
\par
\endgroup
\section{Introduction}
\label{sec:introduction}

4D Gaussian splatting (4DGS)~\cite{li2024spacetime,luiten2024dynamic,wu2024four,yang2024deformable} enables photorealistic dynamic scene reconstruction by representing dynamic scenes with time-varying Gaussian attributes and rendering them through efficient rasterization~\cite{kerbl2023gs}.
These reconstructed scenes can be interacted with on edge devices and allow people to relive their memories~\cite {li2025gifstream, wang2024v3}.
In mainstream 4DGS, a deformation network transforms Gaussians into their states at a given timestamp.
Then, the resulting Gaussians are projected and rendered to a targeted viewpoint.

\textbf{Bottleneck.}
However, evaluating the deformation network for all Gaussians at every frame is costly on edge devices.
For E-D3DGS~\cite{bae2024ed3dgs} on Jetson AGX Orin, deformation accounts for about 85\% of GPU pipeline time on N3DV~\cite{li2022neuralvideo}.
Existing methods reduce computation through Gaussian pruning~\cite{tu2026speede}, static--dynamic decomposition~\cite{chen2025dash,lee2024explicit}, or shared motion representations~\cite{huang2024scgs,kratimenos2024dynmf,kwak2025modec}.
Nevertheless, reducing the Gaussian set or restricting motion expressiveness can compromise reconstruction quality. 
Static–dynamic partitions also remain fixed during rendering, and misclassifying dynamic Gaussians as static can degrade motion reconstruction. 
Our approach retains the pretrained Gaussian set and deformation model, combining temporal prediction with adaptive selection of deformation updates.

\textbf{Insight.}
There is a long history of leveraging short-term temporal regularity in various video-based applications~\cite{bewley2016sort,liu2017voxelflow}.
All these approaches use recent history states to predict future states.
Our key insight here is that many Gaussian trajectories exhibit local temporal regularity (\Fig{fig:method_regularity}).
This motivates us to leverage temporal extrapolation from recent history.
However, prediction reliability varies across Gaussians and over time, and the impact of a state error on the rendered image depends on visibility and rendering contribution.
These observations motivate selecting Gaussians for exact updates based on prediction reliability and rendering contribution.

\textbf{Method.}
Based on our insight, we propose \proj, a post-training acceleration framework (\Sect{sec:methodology}) that enables sparse Gaussian deformation while keeping the pretrained 4DGS model intact without pruning.
During inference, a lightweight router periodically selects a subset of Gaussians for exact deformation updates.
For the remaining Gaussians, we predict their states from the three most recent accepted states.
We merge the exact deformation outputs for the selected Gaussians with the predicted states of the remaining Gaussians for rendering.
In addition, to improve the rendering efficiency, we further reuse the indices of the Gaussian subset across nearby frames to reduce routing overhead, while deformation is recomputed for the selected Gaussians at each frame.

To train the router, we first calibrate a scene-specific update budget using prediction errors collected offline over the calibration frames (\Sect{sec:method_training}).
We then train the router to estimate the update priority of each Gaussian using supervised learning.
Lastly, we validate the budget through offline closed-loop rendering checks and increase it if the quality criteria are not met.

To control accumulated prediction error, \proj{} monitors online drift signals and schedules three consecutive full-deformation frames to rebuild the prediction history (\Sect{sec:method_refresh}).
To mitigate the latency spikes caused by these full refreshes, we use refresh precomputation to distribute part of the deformation workload across preceding frames.

\textbf{Results.}
We evaluate \proj over three datasets~\cite{sabater2017dataset, li2022neuralvideo, yan2023nerfds} using four 4DGS algorithms, E-D3DGS~\cite{bae2024ed3dgs}, 4DGaussians~\cite{wu2024four}, GaGS~\cite{lu2024gags}, and SpeeDe3DGS~\cite{tu2026speede}.
Compared with the corresponding GPU baselines, \proj achieves up to $6.7\times$ speedups on Jetson AGX Orin and up to $8.2\times$ on Orin NX, with dataset-average PSNR decreases of at most $0.16$~dB across the twelve backbone--dataset pairs.
Our contributions are summarized as follows:
\begin{itemize}
\item A sparse deformation framework that combines historical state prediction with selective updates from frozen deformation models.
\item Scene-specific budget calibration and teacher-supervised routing to allocate updates according to prediction reliability and rendering contribution.
\item Drift-aware history refresh and refresh precomputation to control accumulated error and distribute refresh work across frames.
\end{itemize}

\section{Related Work}
\label{sec:related_work}

\textbf{Dynamic Gaussian Representations.}
Canonical deformation fields used in dynamic NeRFs~\cite{park2021nerfies,park2021hypernerf,pumarola2021dnerf} also model Gaussian motion~\cite{das2024npg}, while other representations explicitly model spacetime evolution~\cite{duan2024rotor,wang2025freetime,yang2024native4d}.
Deformable 3D Gaussians~\cite{yang2024deformable} maps canonical primitives to time-varying states, while 4DGaussians~\cite{wu2024four} combines factorized 4D features~\cite{cao2023hexplane,fridovich2023kplanes} with a lightweight deformation decoder.
GaGS~\cite{lu2024gags} incorporates geometry features, and E-D3DGS~\cite{bae2024ed3dgs} combines per-Gaussian and temporal embeddings with coarse and fine deformations.
Explicit temporal approaches~\cite{katsumata2024compact,lin2024flow,park2025spline} include Spacetime Gaussian Feature Splatting~\cite{li2024spacetime}, which combines temporal opacity, parametric motion and rotation, and feature rendering.
For deformation-based models, repeated network evaluations remain costly.
\proj{} learns to allocate evaluations for sequential rendering with frozen pretrained parameters.

\textbf{Reducing Dynamic Computation.}
Compact motion models~\cite{kratimenos2024dynmf,zhao2024prediction} reduce computation, while pruning and compact encoding improve efficiency for static~\cite{fan2024lightgaussian,fang2024mini,girish2024eagles} and dynamic scenes~\cite{zhang2025mega}.
SC-GS~\cite{huang2024scgs} interpolates motion from sparse control points, while MoDec-GS~\cite{kwak2025modec} uses global and local scaffolds.
DASH~\cite{chen2025dash} combines static--dynamic decomposition with 4D hash encoding, and SpeeDe3DGS~\cite{tu2026speede} combines temporal pruning and sampling with groupwise rigid motion.
However, pruning or motion constraints can impair fidelity, while fixed static--dynamic partitions do not adapt update allocation over time.
\proj{} retains the pretrained Gaussian set and deformation model, periodically selecting Gaussians for exact updates using learned priorities and predicting remaining states from history.

\textbf{Temporal Reuse and Selective Computation.}
Temporal coherence supports streaming reconstruction~\cite{gao2024hicom,girish2024queen,hu2025gc,yan2025instant} and long sequences~\cite{xu2026layered,xu2024hierarchy}.
3DGStream~\cite{sun2024stream} uses a compact neural transformation cache, while SWinGS~\cite{shaw2024swings} fits consistent representations over sliding windows.
Both exploit temporal structure during fitting.
At inference, CaT-GS~\cite{zhang2026catgs} reuses culling and sorting results and schedules tiles to balance rasterization.
Video methods reuse intermediate states through sparse updates~\cite{dutson2022event,parger2022deltacnn,parger2023motiondelta}, and Eventful Transformers~\cite{dutson2023eventful} selectively recomputes tokens.
\proj{} applies selective computation to frozen Gaussian deformation models, predicting states from accepted history and learning which Gaussians require exact updates.
Because predicted states feed later predictions, three-frame history refreshes control accumulated error, while refresh precomputation distributes work across frames.

\section{Motivation}

\label{sec:method_motivation}

\begin{figure}[!tbp]
\centering
\input{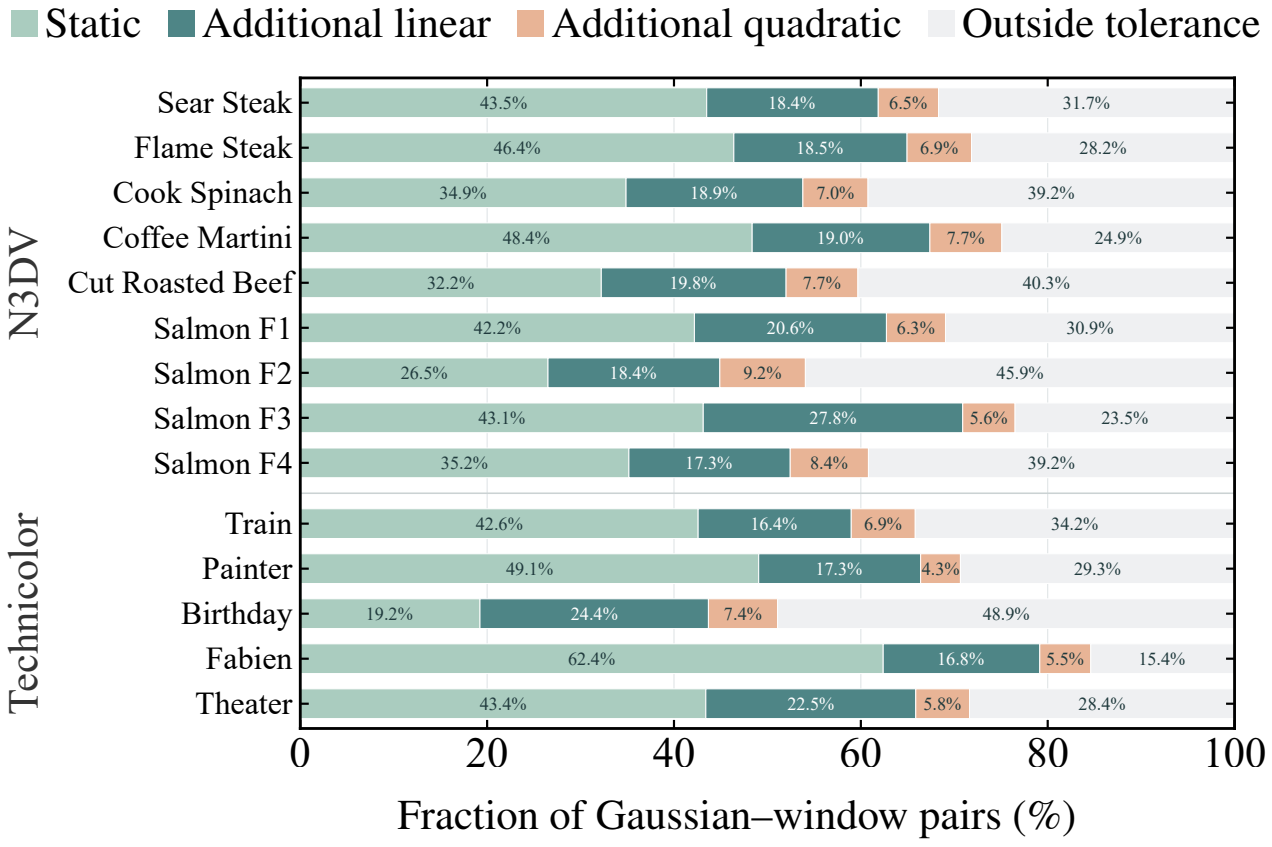}
\caption{Five-frame positional regularity of E-D3DGS on N3DV and Technicolor. Bars partition Gaussian--window pairs into mutually exclusive static, additional linear, additional quadratic, and outside-tolerance cases. Fits use a maximum positional-error tolerance of $0.1$ times the geometric-mean canonical scale.}
\label{fig:method_regularity}
\end{figure}

In this section, we first show that Gaussian center trajectories often exhibit local temporal regularity.
\Fig{fig:method_regularity} analyzes five-frame E-D3DGS~\cite{bae2024ed3dgs} trajectories on two datasets: N3DV~\cite{li2022neuralvideo} and Technicolor~\cite{sabater2017dataset}.
We classify each Gaussian trajectory into: static, linear, quadratic, and other moves.
On average, static, linear, and quadratic fits jointly cover 66.2\% of Gaussian-window pairs on N3DV and 68.8\% on Technicolor.
Additional linear and quadratic fits cover 26.5\% of all pairs.
These results motivate exploiting local temporal regularity for inexpensive history-based prediction.

\begin{figure}[t]
\centering
\input{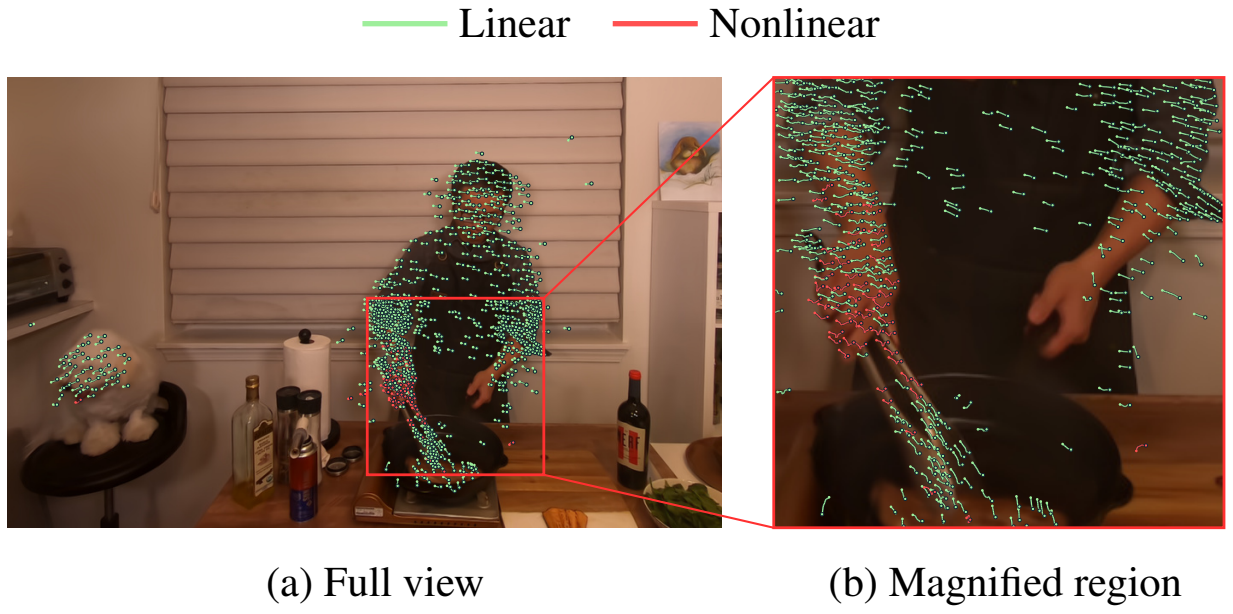}
\caption{Projected Gaussian center trajectories in \textit{sear\_steak}~\cite{li2022neuralvideo} over frames 111--130, overlaid on the frame-130 rendering from a fixed viewpoint. The right panel magnifies the boxed region. Green marks locally linear trajectories; red marks trajectories exceeding a geometric straightness threshold.}
\label{fig:method_trajectories}
\end{figure}

However, this local regularity varies across both Gaussians and time.
\Fig{fig:method_trajectories} shows locally straight and nonlinear projected trajectories co-existing in a \textit{sear steak} scene.
E.g., a red trajectory may be locally straight before a turn, but the change in direction can make extrapolation from recent motion less reliable.
Across N3DV sequences, 54.4\% of sampled Gaussians satisfy the five-frame fitting criterion in some windows but violate it in others.
These observations motivate adapting the selection of Gaussians for exact updates over time.

Choosing which Gaussians to update also requires considering how prediction errors in position, rotation, scale, and appearance affect the rendered image.
We therefore combine history-based prediction with learned update priorities based on each Gaussian's state history and rendering contribution.
To limit error accumulation, we also rebuild the prediction history with full-deformation refreshes.
We evaluate prediction and history refresh in \Sect{sec:ablation}.

\begin{figure*}[t]
\centering
\includegraphics[width=\linewidth]{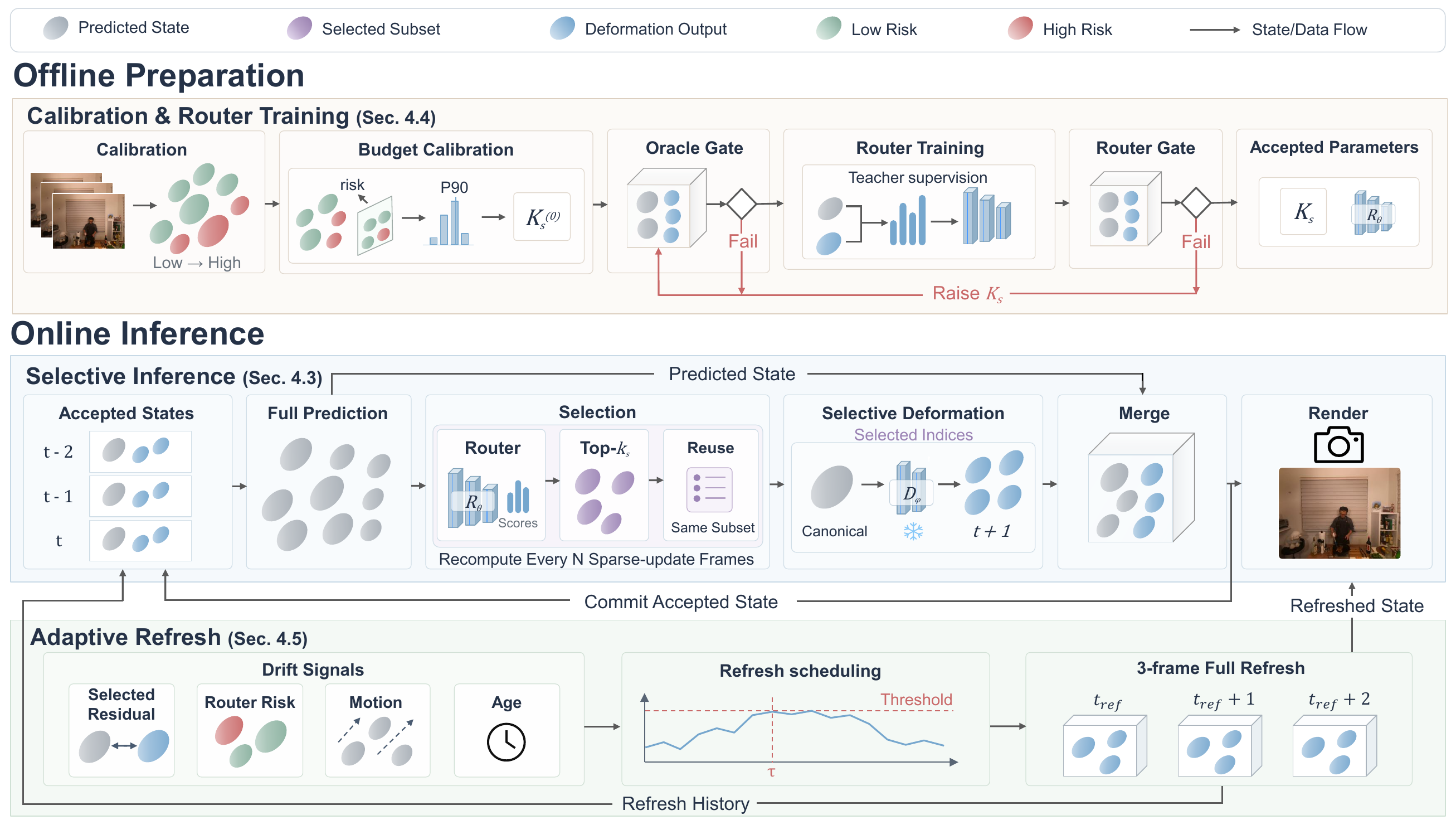}
\caption{Overview of \proj{}. (1) Calibration and training determine the scene budget $K_s$ and router $R_\theta$ through oracle and router quality gates. (2) Selective inference predicts all attributes from accepted history, selects and reuses indices, and replaces selected predictions with exact outputs from frozen $\mathcal D_\phi$. The merged state is rendered and stored in history. (3) Adaptive refresh uses drift signals to schedule three consecutive full-deformation frames that rebuild history, with partial precomputation after target timestamps are fixed.}
\label{fig:method_overview}
\end{figure*}

\vspace{-6pt}
\section{Methodology}

\label{sec:methodology}

Based on the insights in \Sect{sec:method_motivation}, we propose \proj, a post-training framework combining history-based prediction with selective exact deformation to reduce rendering cost.


\subsection{Preliminaries}

\label{sec:method_preliminaries}

Let $G$ denote the number of Gaussians. Deformation-based dynamic Gaussian representations~\cite{yang2024deformable,wu2024four,bae2024ed3dgs} describe a scene using canonical Gaussians $\mathcal{G}^{0}=\{g_i^{0}\}_{i=1}^{G}$, each with a center, rotation, scale, opacity, and spherical-harmonic appearance coefficients~\cite{kerbl2023gs}. A deformation network maps canonical attributes, associated features, and time to time-varying attributes. For compact notation, $g_i^0$ also includes the frozen per-Gaussian inputs required by the backbone, such as learned embeddings; shared deformation parameters are included in $\phi$. We use $\mathcal{D}_{\phi}$ to denote deformation evaluation together with its composition with canonical attributes:
\begin{equation}
\mathbf{y}_{i,t}^{\star}
=\mathcal{D}_{\phi}(g_i^{0},t)
=(\boldsymbol\mu_{i,t},\mathbf q_{i,t},\mathbf s_{i,t},o_{i,t},\mathbf h_{i,t}).
\label{eq:deformation_state}
\end{equation}
The renderer projects the Gaussian states and performs standard front-to-back alpha compositing~\cite{kerbl2023gs}. A backbone may deform only a subset of the attributes. \proj{} freezes canonical Gaussian and deformation parameters, trains only the router, and renders all Gaussians. We target sequential rendering at uniform time steps; an exact update evaluates the deformation model at the target timestamp.

\subsection{Overview}

\label{sec:method_overview}

\Fig{fig:method_overview} gives an overview of \proj, which consists of offline preparation and online inference.
Offline, we calibrate a scene-specific exact-update percentage budget, $K_s$, to determine how many Gaussians can receive exact updates.
Then, we train a lightweight router to select Gaussians under this fixed budget (\Sect{sec:method_training}).
Online, we predict all Gaussian states using recent history and select which Gaussians should receive exact deformation updates.
Lastly, we merge the predicted states of unselected Gaussians and the exact deformation outputs of selected Gaussians together for rendering (\Sect{sec:method_sparse}).
To reduce routing overhead, we reselect Gaussians every $N=5$ frames and reuse their indices in between.
The selected Gaussians still receive exact deformation updates at every sparse-update frame.
In addition, to control accumulated prediction error, we use online drift signals to schedule three consecutive full-deformation frames that rebuild the history (\Sect{sec:method_refresh}).

\subsection{Predictive Sparse Deformation}
\label{sec:method_sparse}

To reduce deformation cost, we combine history-based prediction of all Gaussian states with exact deformation updates for a selected subset.

\textbf{Full Prediction.}
As shown in \Fig{fig:method_overview}, we first predict all Gaussian attributes so that unselected Gaussians can be updated without evaluating the deformation network.
Let $\mathbf Y_t=\{\mathbf y_{i,t}\}_{i=1}^{G}$ denote the complete state rendered at frame $t$.
Three full-deformation frames initialize the history.

\begingroup
\predisplaypenalty=0
We then predict $\widehat{\mathbf Y}_{t+1}=\mathcal P(\mathbf Y_{t-2},\mathbf Y_{t-1},\mathbf Y_t)$, using damped finite-difference extrapolation for positions:
\begin{equation}
\begin{aligned}
\mathbf v_{i,t}&=\boldsymbol\mu_{i,t}-\boldsymbol\mu_{i,t-1},\\
\mathbf a_{i,t}&=\boldsymbol\mu_{i,t}-2\boldsymbol\mu_{i,t-1}+\boldsymbol\mu_{i,t-2},\\
\widehat{\boldsymbol\mu}_{i,t+1}&=\boldsymbol\mu_{i,t}
+\lambda_v\mathbf v_{i,t}+\lambda_a\mathbf a_{i,t}.
\end{aligned}
\label{eq:history_prediction}
\end{equation}
\endgroup
We use $\lambda_v=0.85$ and $\lambda_a=0.15$.
Rotations use damped quaternion increments, scales retain their latest values, and dynamic appearance attributes are filtered from the same three states.
Static attributes retain their canonical values.

\textbf{Learned Selection and Reuse.}
Selection accounts for prediction error and rendering contribution.
Measuring current errors for all Gaussians requires full deformation, so we train a lightweight MLP router $R_\theta$ with full-deformation supervision offline (\Sect{sec:method_training}).
It outputs an update score $\ell_{i,t+1}$ from a frozen Gaussian descriptor, an encoded timestamp, and online features describing motion, visibility, contribution, and update age.
The online features are computed from accepted history and the current camera, with previously observed prediction residuals included as additional inputs.

\begingroup
\predisplaypenalty=0
To reduce routing overhead, we recompute features, scores, and selected indices every $N=5$ sparse-update frames (\Sect{sec:sensitivity}) and reuse the indices in between.
Periodic reselection follows changes in motion, visibility, and contribution.
Let $\mathcal T_R$ denote recomputation frames and $\mathcal S_{\mathrm{cache}}$ the latest valid selection:
\begin{equation}
\mathcal S_{t+1}=\begin{cases}
\operatorname{TopK}_{k_s}\!\left(\{\ell_{i,t+1}\}_{i=1}^{G}\right),
&t+1\in\mathcal T_R,\\
\mathcal S_{\mathrm{cache}},&\text{otherwise}.
\end{cases}
\label{eq:router_selection}
\end{equation}
\endgroup
Here, $\operatorname{TopK}_{k_s}$ returns the indices of the $k_s$ largest scores, where $k_s=\lceil K_sG\rceil$ remains fixed during deployment.
At each sparse-update frame, selected Gaussians receive exact updates, while the others retain their predictions.
A history refresh invalidates the selection cache (\Sect{sec:method_refresh}).

\textbf{State Merging and Feedback.}
On each router recomputation frame $u$, we measure the Euclidean distance $\epsilon_{i,u}$ between selected Gaussians' predicted and exact positions before merging.
We cache these raw errors separately for refresh scheduling (\Sect{sec:method_refresh}).
Dividing them by their selected-set mean plus $10^{-6}$ gives $\widetilde\epsilon_{i,u}$, stored as pointwise routing feedback; unmeasured entries retain their previous values.

At every sparse-update frame, we replace selected predictions with exact outputs to form $\mathbf Y_{t+1}$, then rasterize and store it in history.
Both residual caches remain unchanged between router recomputations.

\subsection{Budget Calibration and Router Training}

\label{sec:method_training}

To preserve quality with limited Gaussian updates via deformation inference, we calibrate a budget for each scene offline and train a router to select important Gaussians online.

\textbf{Budget Calibration.}
We calibrate a scene-specific exact-update budget offline using prediction errors and rendering contribution.
Comparing predicted and full-deformation states gives the rendering-aware risk:
\begin{equation}
\rho_{i,t}=\frac{m^{\mathrm{vis}}_{i,t}o_{i,t}\pi r_{i,t}^{2}}{HW}
\left[\log(1+e_{i,t})\right]^2.
\label{eq:calibration_risk}
\end{equation}
Here, $e_{i,t}$ combines position, rotation, scale, opacity, and SH prediction errors with projected-footprint weighting.
The visibility mask $m^{\mathrm{vis}}_{i,t}$, activated opacity $o_{i,t}$, and projected radius $r_{i,t}$ come from full-model renderings of size $H\times W$.

We derive a frame-wise budget proposal from the total risk and its spatial spread, with attenuation at low risk.
We take the 90th percentile of these proposals over calibration frames and round it up to a predefined budget tier, $K_s^{(0)}$.
Because prediction errors can accumulate over time, we first test this budget in an oracle-driven closed-loop rollout.

\textbf{Router Training.}
Once the oracle check passes, we collect teacher supervision through offline closed-loop rollouts.
The teacher combines the five attribute errors after normalizing each by its frame-wise median.
Visibility, predicted opacity, and projected radius then weight the combined error by rendering contribution.
The top-$k_s$ scores define binary update labels $z_i$, while a bounded transform of the scores gives continuous priorities $u_i\in[0,1]$.

\begingroup
\predisplaypenalty=0
Because the router selects a fixed-size subset, we emphasize pairs near the selection boundary.
Each $(i,j)\in\mathcal P_b$ contains a teacher-selected Gaussian $i$ and an unselected Gaussian $j$.
We optimize
\begin{equation}
\begin{aligned}
\mathcal L_{\mathrm{rank}}&=\mathbb E_{(i,j)\in\mathcal P_b}
\log\!\left(1+\exp\!\left[-\frac{\ell_i-\ell_j}{T}\right]\right),\\
\mathcal L&=\mathcal L_{\mathrm{rank}}
+\lambda_c\,\mathbb E_i\operatorname{BCE}(\sigma(\ell_i),z_i)\\
&\quad+\lambda_s\,\mathbb E_i\operatorname{SmoothL1}(\sigma(\ell_i),u_i).
\end{aligned}
\label{eq:router_training}
\end{equation}
\endgroup
Here, $\sigma$ is the sigmoid function and $T$ is the ranking temperature.
The three terms supervise boundary ordering, update membership, and continuous priority, respectively.
We train a router for each scene--backbone--budget combination.

\textbf{Closed-Loop Validation.}
We then validate the trained router in a second closed-loop rollout. 
Both checks compare sparse and full-model renderings, requiring a mean proxy PSNR of at least 40 dB and a 10th percentile of at least 37 dB.
If either check fails, we raise the budget and repeat the procedure at the new tier.
If no sparse tier passes, we use full deformation.
The accepted $K_s$ remains fixed during deployment.
Budget proposal formulas, error definitions, proxy-image construction, and training details are provided in Supplementary Sec.~B.

\subsection{Adaptive Refresh and Amortized Execution}

\label{sec:method_refresh}

To limit accumulated prediction error, we adaptively refresh the state history through full deformation and amortize part of the refresh computation across preceding frames.

\textbf{Drift Monitoring.}
Predicted states feed later predictions, so their errors can accumulate.
We monitor this drift to schedule three consecutive full-deformation frames that rebuild history.
At each sparse-update frame, we compute
\begin{equation}
d_t=0.4\bar E_t+0.2\bar r_t+0.2\bar m_t+0.2a_t.
\label{eq:drift_score}
\end{equation}
Here, $E_t=|\mathcal S_u|^{-1}\sum_{i\in\mathcal S_u}\epsilon_{i,u}$ averages the independently cached raw pre-merge positional errors at the latest router recomputation $u$ (\Sect{sec:method_sparse}), retained between recomputations.
The terms $\bar E_t$, $\bar r_t$, and $\bar m_t$ are normalized increases in positional error, router score, and motion above their post-refresh baselines, clipped to $[0,1]$.
The age term $a_t$ is the fraction of currently visible Gaussians whose last exact update was at least 20 frames ago.
The score uses current visibility and age together with cached router outputs and $E_t$.

\textbf{Refresh Scheduling.}
After a minimum interval of 20 frames, two consecutive frame-level scores of at least $0.6$ trigger scheduling at the next frame.
Scheduling is forced by $s+40$, where $s$ is the start of the previous full refresh.
If scheduling begins at $\tau$, the first refresh target is $t_{\mathrm{ref}}=\tau+10$, followed by $t_{\mathrm{ref}}+1$ and $t_{\mathrm{ref}}+2$.

\textbf{Refresh Precomputation.}
The evaluated backbones compute deformation from canonical attributes, frozen features, and the target timestamp.
Once the targets are fixed, we distribute exact evaluations for a budget-dependent Gaussian subset over the ten frames before the first target, caching its state separately for each target.

At each target frame, we combine the corresponding cached states with exact outputs for the remaining Gaussians to form a full state.
View-dependent SH evaluation and rasterization use that frame's camera, and the full state is committed to history.
These three consecutive full states rebuild all three entries in the prediction history used by the predictor in \Sect{sec:method_sparse}, and the refresh invalidates the selection cache.
Supplementary Sec.~C details signal normalization and refresh implementation.

\section{Evaluation}
\label{sec:experiments}

\begin{table*}[!t]
\centering
\caption{Dataset-average quality and efficiency of four frozen backbones with and without \proj{}. FPS and FPS P5 are measured on Jetson AGX Orin. $\#\mathrm{Inst.}$ denotes millions of GPU warp instructions per frame from separate server-side profiling.}
\label{tab:main_results}
\footnotesize
\renewcommand{\arraystretch}{0.93}
\setlength{\tabcolsep}{2pt}
\setlength{\aboverulesep}{1.5pt}
\setlength{\belowrulesep}{2pt}
\begin{tabular*}{\textwidth}{@{\extracolsep{\fill}}clccccccccc@{}}
\toprule
\multirow{2}{*}{Dataset} & \multirow{2}{*}{Method} & \multicolumn{3}{c}{Rendering Quality} & \multicolumn{2}{c}{Temporal Quality} & \multicolumn{4}{c}{Efficiency} \\
\cmidrule(lr){3-5}\cmidrule(lr){6-7}\cmidrule(lr){8-11}
& & PSNR$\uparrow$ & SSIM$\uparrow$ & LPIPS$\downarrow$ & tOF$\downarrow$ & tLP$\times100\downarrow$ & FPS$\uparrow$ & Speedup$\uparrow$ & FPS P5$\uparrow$ & \#Inst.\,$(10^6)\downarrow$ \\
\midrule
\multirow{8}{*}{\rotatebox[origin=c]{90}{\textbf{Technicolor}~\cite{sabater2017dataset}}} & E-D3DGS~\cite{bae2024ed3dgs} & 34.07 & 0.937 & 0.0783 & 0.127 & 0.557 & 21.97 & -- & 21.87 & 2077.66 \\
& \textbf{E-D3DGS + \proj{}} & 33.91 & 0.936 & 0.0789 & 0.128 & 0.555 & \effvalue{72.78} & \effvalue{3.31$\times$} & \effvalue{64.09} & \effvalue{672.37} \\
\arrayrulecolor{tableseparator}\cmidrule[0.3pt](l){2-11}\arrayrulecolor{black}
& 4DGaussians~\cite{wu2024four} & 32.32 & 0.906 & 0.1062 & 0.172 & 0.596 & 21.68 & -- & 21.64 & 1564.12 \\
& \textbf{4DGaussians + \proj{}} & 32.27 & 0.905 & 0.1063 & 0.176 & 0.571 & \effvalue{59.43} & \effvalue{2.74$\times$} & \effvalue{46.44} & \effvalue{556.57} \\
\arrayrulecolor{tableseparator}\cmidrule[0.3pt](l){2-11}\arrayrulecolor{black}
& GaGS~\cite{lu2024gags} & 32.99 & 0.927 & 0.0957 & 0.156 & 0.425 & 10.65 & -- & 10.63 & 8206.42 \\
& \textbf{GaGS + \proj{}} & 32.92 & 0.927 & 0.0959 & 0.156 & 0.426 & \effvalue{71.79} & \effvalue{6.74$\times$} & \effvalue{52.58} & \effvalue{760.01} \\
\arrayrulecolor{tableseparator}\cmidrule[0.3pt](l){2-11}\arrayrulecolor{black}
& SpeeDe3DGS$^{\dagger}$~\cite{tu2026speede} & 30.95 & 0.901 & 0.0834 & 0.103 & 0.413 & 8.03 & -- & 8.01 & 6462.78 \\
& \textbf{SpeeDe3DGS$^{\dagger}$ + \proj{}} & 30.91 & 0.900 & 0.0844 & 0.103 & 0.415 & \effvalue{37.50} & \effvalue{4.67$\times$} & \effvalue{30.43} & \effvalue{1716.63} \\
\midrule
\multirow{8}{*}{\rotatebox[origin=c]{90}{\textbf{N3DV}~\cite{li2022neuralvideo}}} & E-D3DGS~\cite{bae2024ed3dgs} & 27.92 & 0.927 & 0.1139 & 0.151 & 1.623 & 15.64 & -- & 15.59 & 2517.13 \\
& \textbf{E-D3DGS + \proj{}} & 27.88 & 0.927 & 0.1144 & 0.152 & 1.622 & \effvalue{38.24} & \effvalue{2.45$\times$} & \effvalue{20.66} & \effvalue{1036.18} \\
\arrayrulecolor{tableseparator}\cmidrule[0.3pt](l){2-11}\arrayrulecolor{black}
& 4DGaussians~\cite{wu2024four} & 27.60 & 0.919 & 0.1259 & 0.159 & 1.633 & 29.59 & -- & 29.52 & 1132.30 \\
& \textbf{4DGaussians + \proj{}} & 27.57 & 0.919 & 0.1263 & 0.160 & 1.630 & \effvalue{61.21} & \effvalue{2.07$\times$} & \effvalue{42.92} & \effvalue{680.32} \\
\arrayrulecolor{tableseparator}\cmidrule[0.3pt](l){2-11}\arrayrulecolor{black}
& GaGS~\cite{lu2024gags} & 23.52 & 0.902 & 0.2193 & 0.215 & 3.611 & 9.59 & -- & 9.56 & 4499.06 \\
& \textbf{GaGS + \proj{}} & 23.49 & 0.902 & 0.2198 & 0.216 & 3.604 & \effvalue{30.14} & \effvalue{3.14$\times$} & \effvalue{18.63} & \effvalue{1555.07} \\
\arrayrulecolor{tableseparator}\cmidrule[0.3pt](l){2-11}\arrayrulecolor{black}
& SpeeDe3DGS$^{\dagger}$~\cite{tu2026speede} & 23.92 & 0.884 & 0.2459 & 0.221 & 3.706 & 15.77 & -- & 15.70 & 3279.77 \\
& \textbf{SpeeDe3DGS$^{\dagger}$ + \proj{}} & 23.91 & 0.883 & 0.2461 & 0.222 & 3.702 & \effvalue{33.54} & \effvalue{2.13$\times$} & \effvalue{22.27} & \effvalue{1445.12} \\
\midrule
\multirow{8}{*}{\rotatebox[origin=c]{90}{\textbf{NeRF-DS}~\cite{yan2023nerfds}}} & E-D3DGS~\cite{bae2024ed3dgs} & 23.78 & 0.841 & 0.0884 & 0.274 & 0.487 & 62.26 & -- & 61.79 & 445.14 \\
& \textbf{E-D3DGS + \proj{}} & 23.77 & 0.841 & 0.0890 & 0.277 & 0.483 & \effvalue{113.44} & \effvalue{1.82$\times$} & \effvalue{64.03} & \effvalue{222.19} \\
\arrayrulecolor{tableseparator}\cmidrule[0.3pt](l){2-11}\arrayrulecolor{black}
& 4DGaussians~\cite{wu2024four} & 22.59 & 0.803 & 0.1322 & 0.410 & 0.782 & 72.26 & -- & 71.75 & 390.68 \\
& \textbf{4DGaussians + \proj{}} & 22.58 & 0.803 & 0.1323 & 0.411 & 0.783 & \effvalue{126.27} & \effvalue{1.75$\times$} & \effvalue{76.55} & \effvalue{273.86} \\
\arrayrulecolor{tableseparator}\cmidrule[0.3pt](l){2-11}\arrayrulecolor{black}
& GaGS~\cite{lu2024gags} & 23.53 & 0.835 & 0.1155 & 0.360 & 0.636 & 29.76 & -- & 29.59 & 1368.96 \\
& \textbf{GaGS + \proj{}} & 23.51 & 0.834 & 0.1171 & 0.367 & 0.662 & \effvalue{99.82} & \effvalue{3.35$\times$} & \effvalue{65.79} & \effvalue{379.39} \\
\arrayrulecolor{tableseparator}\cmidrule[0.3pt](l){2-11}\arrayrulecolor{black}
& SpeeDe3DGS$^{\dagger}$~\cite{tu2026speede} & 23.78 & 0.853 & 0.0999 & 0.280 & 0.506 & 42.14 & -- & 41.81 & 927.16 \\
& \textbf{SpeeDe3DGS$^{\dagger}$ + \proj{}} & 23.68 & 0.848 & 0.1039 & 0.290 & 0.515 & \effvalue{83.59} & \effvalue{1.98$\times$} & \effvalue{44.13} & \effvalue{452.83} \\
\bottomrule
\end{tabular*}
\par\vspace{1pt}
\begin{minipage}{\textwidth}
\footnotesize\raggedright
$^{\dagger}$ SpeeDe3DGS uses TSP and TSS, without GroupFlow.
\end{minipage}
\end{table*}

\begin{figure*}[!t]
\centering
\resizebox{0.49\textwidth}{!}{%
\begin{overpic}[width=360bp]{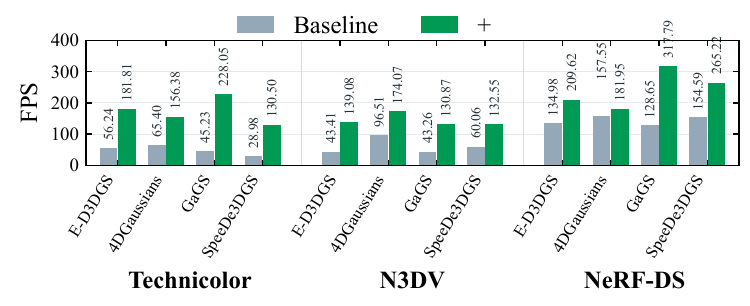}
\put(66.31333,36.21266){\normalfont\fontsize{11.82948bp}{14.19538bp}\selectfont\proj{}}
\end{overpic}}\hfill
\resizebox{0.49\textwidth}{!}{%
\begin{overpic}[width=360bp]{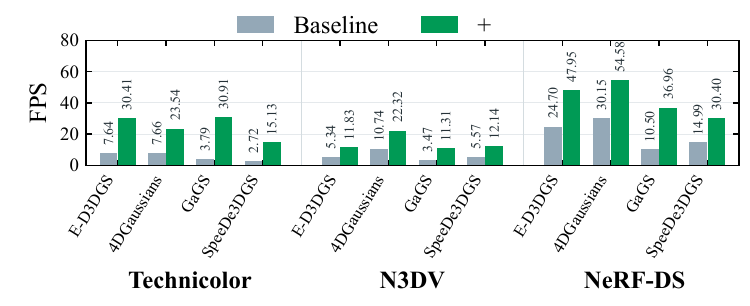}
\put(66.31333,36.21266){\normalfont\fontsize{11.82948bp}{14.19538bp}\selectfont\proj{}}
\end{overpic}}\par
\vspace{3pt}
{\normalfont\fontsize{8}{9.6}\selectfont
\noindent\makebox[0.49\textwidth][c]{(a) A6000}\hfill
\makebox[0.49\textwidth][c]{(b) Orin NX}\par}
\caption{Dataset-average throughput on (a) A6000 and (b) Orin NX. Gray bars show the frozen-backbone baselines, and green bars show \proj{}. All twelve backbone--dataset pairs benefit, with peak speedups of 5.04$\times$ on A6000 and 8.15$\times$ on Orin NX.}
\label{fig:cross_device_fps}
\end{figure*}

\begin{table}[t]
\centering
\caption{Comparison with DASH~\cite{chen2025dash} on Technicolor~\cite{sabater2017dataset}: five-scene means at $1024\times544$. DASH-style is our fixed-partition control; DASH is reproduced with official code.}
\label{tab:dash_comparison}

\small
\setlength{\tabcolsep}{3pt}
\begin{tabular*}{\columnwidth}{@{\extracolsep{\fill}}llcc@{}}
\toprule
Backbone & Configuration & PSNR$\uparrow$ & FPS$\uparrow$ \\
\midrule
\multirow{3}{*}{E-D3DGS~\cite{bae2024ed3dgs}} & Baseline & 34.07 & 21.97 \\
 & DASH-style & 33.54 & 72.97 \\
 & \textbf{Ours} & 33.91 & 72.78 \\
\methodrule
\multirow{3}{*}{4DGaussians~\cite{wu2024four}} & Baseline & 32.32 & 21.68 \\
 & DASH-style & 32.01 & 59.69 \\
 & \textbf{Ours} & 32.27 & 59.43 \\
\methodrule
\multirow{3}{*}{GaGS~\cite{lu2024gags}} & Baseline & 32.99 & 10.65 \\
 & DASH-style & 32.70 & 39.60 \\
 & \textbf{Ours} & 32.92 & 71.79 \\
\methodrule
\multirow{3}{*}{SpeeDe3DGS~\cite{tu2026speede}} & Baseline & 30.95 & 8.03 \\
 & DASH-style & 29.72 & 29.68 \\
 & \textbf{Ours} & 30.91 & 37.50 \\
\midrule
\multicolumn{2}{l}{DASH~\cite{chen2025dash}} & 31.12 & 39.13 \\
\bottomrule
\end{tabular*}
\end{table}

\subsection{Experimental Setup}

We evaluate \proj{} on Technicolor~\cite{sabater2017dataset}, N3DV~\cite{li2022neuralvideo}, and NeRF-DS~\cite{yan2023nerfds}, comprising 5, 9, and 7 sequences, respectively, using E-D3DGS~\cite{bae2024ed3dgs}, 4DGaussians~\cite{wu2024four}, GaGS~\cite{lu2024gags}, and SpeeDe3DGS~\cite{tu2026speede}. Each backbone and its \proj{} variant share the same pretrained checkpoint and gsplat/AccuTile renderer~\cite{ye2025gsplat}. All SpeeDe3DGS experiments use TSP and TSS, without GroupFlow. Unless otherwise stated, we use a selection-reuse period of \(N=5\). For Technicolor's 50-frame sequences, we disable refreshes after initialization. Throughput is measured on a Jetson AGX Orin 64 GB Developer Kit, an Orin NX 16 GB, and an A6000, with both Jetsons operating in MAXN mode.

We evaluate quality over full sequences using PSNR, SSIM~\cite{wang2004ssim}, AlexNet LPIPS~\cite{zhang2018lpips}, and the temporal metrics tOF and $\mathrm{tLP}\times100$~\cite{chu2020tecogan}. FPS is the number of rendered frames divided by synchronized wall-clock time, including prediction, selection, deformation, rendering, and refresh computation. Backbone timing and instruction profiling exclude the three initialization frames. We time each experiment independently over three runs per scene and average scenes equally. Speedup is the ratio of dataset-average FPS. FPS P5 is the fifth percentile of instantaneous frame rates, computed per run and then averaged across runs and scenes. Higher values indicate better slow-frame performance. We also report executed GPU warp instructions per frame.

Scene-specific preparation uses training views and timestamps and takes 1–2 minutes on an RTX A6000, plus about one minute for budget escalation after failed validation.
Routers have about 2.0k--2.6k parameters. 
Implementation and measurements are shown in Supplementary Sec.~D.1--Sec.~D.3 and Sec.~E.

\subsection{Main Results}

\Tbl{tab:main_results} compares image quality, temporal consistency, and rendering efficiency across twelve backbone–dataset pairs. \proj{} achieves 1.75--6.74$\times$ speedups on Jetson AGX Orin, with dataset-average PSNR decreases of at most 0.16 dB. SSIM, LPIPS, and the temporal metrics remain close to their full-deformation baselines, indicating that the speedups largely preserve image and temporal quality.

\textbf{Slow-frame Performance.}
FPS P5 improves across all twelve pairs. For GaGS on Technicolor, it increases from 10.63 to 52.58 FPS. \proj{} therefore improves both average throughput and the frame rates experienced during slower parts of rendering.

\textbf{Executed Computation.}
Separate server-side profiling shows a 29.9\%--90.7\% reduction in executed GPU warp instructions per frame. GaGS on Technicolor decreases from 8206.42 to 760.01 million instructions per frame. These measurements confirm that sparse deformation reduces the computational work of the rendering pipeline.

\textbf{Results on Other Devices.}
\Fig{fig:cross_device_fps} shows throughput gains across all twelve pairs on both A6000 and Orin NX, reaching 5.04$\times$ on A6000 and 1.81--8.15$\times$ on Orin NX. GaGS on Technicolor achieves the largest gain on both devices, increasing from 45.23 to 228.05 FPS on A6000 and from 3.79 to 30.91 FPS on Orin NX.

\textbf{Camera Motion.}
With unchanged routers and budgets, E-D3DGS + \proj{} achieves 48.54 dB PSNR against full-deformation renderings along moving-camera trajectories across all nine N3DV scenes (Supplementary Sec.~D.10).

\subsection{Comparison with DASH}

DASH~\cite{chen2025dash} fixes its static--dynamic partition for the entire sequence, while \proj{} periodically reselects Gaussians for exact updates. On Technicolor, we compare DASH-style controls on pretrained backbones and a separately trained reproduction of the official DASH implementation.

Against DASH-style, \proj{} improves mean PSNR for all four backbones (\Tbl{tab:dash_comparison}). E-D3DGS and 4DGaussians gain 0.37 and 0.26 dB, respectively, with less than 0.5\% lower FPS. GaGS and SpeeDe3DGS gain 0.22 and 1.19 dB, with 81.3\% and 26.3\% higher FPS, respectively. At matched scene budgets, \proj{} improves mean PSNR over a fixed-partition variant using the same history-based prediction by 0.26--1.69 dB across the four backbones (Supplementary Tbl.~S4).
\subsection{Ablation Studies}
\label{sec:ablation}

\Tbl{tab:ablation_technicolor} evaluates learned routing, motion prediction, and scene-specific budgets on 50-frame Technicolor sequences. \Tbl{tab:ablation_n5} uses 300-frame N3DV sequences to evaluate history refresh over longer rendering sequences. Supplementary Sec.~D.4--Sec.~D.5 describe the experimental protocols.

\begin{table}[!tbp]
\centering
\caption{Module ablations on 50-frame Technicolor~\cite{sabater2017dataset} sequences, averaged equally over five scenes and four backbones. FPS is measured on Jetson AGX Orin.}
\label{tab:ablation_technicolor}
\small
\setlength{\tabcolsep}{2pt}
\begin{tabular*}{\columnwidth}{@{\extracolsep{\fill}}p{83pt}cccc@{}}
\toprule
Variant & PSNR$\uparrow$ & SSIM$\uparrow$ & LPIPS$\downarrow$ & FPS$\uparrow$ \\
\midrule
Baseline & 32.58 & 0.918 & 0.0909 & 15.50 \\
\methodrule
Motion-only selection & 31.06 & 0.906 & 0.1018 & 62.50 \\
Hold-last prediction & 32.29 & 0.916 & 0.0919 & 60.84 \\
Fixed $K_s=20\%$ & 31.29 & 0.908 & 0.0969 & 67.57 \\
\methodrule
\textbf{Ours} & 32.50 & 0.917 & 0.0914 & 60.47 \\
\bottomrule
\end{tabular*}
\end{table}

\begin{table}[!tbp]
\centering
\caption{History refresh and precomputation ablations on 300-frame N3DV~\cite{li2022neuralvideo} sequences, averaged equally over nine scenes and four backbones. No precomp. preserves the refresh schedule and target frames. No refresh disables both refresh and precomputation. FPS and FPS P5 are measured on Jetson AGX Orin.}
\label{tab:ablation_n5}
\small
\setlength{\tabcolsep}{2pt}
\begin{tabular*}{\columnwidth}{@{\extracolsep{\fill}}lccccc@{}}
\toprule
Variant & PSNR$\uparrow$ & SSIM$\uparrow$ & LPIPS$\downarrow$ & FPS$\uparrow$ & FPS P5$\uparrow$ \\
\midrule
Baseline & 25.74 & 0.908 & 0.1763 & 17.67 & 17.62 \\
\methodrule
No precomp. & 25.71 & 0.908 & 0.1767 & 40.95 & 17.68 \\
No refresh & 25.50 & 0.905 & 0.1803 & 46.93 & 40.44 \\
\methodrule
\textbf{Ours} & 25.71 & 0.908 & 0.1767 & 40.81 & 26.20 \\
\bottomrule
\end{tabular*}
\end{table}

\textbf{Learned Routing.}
We replace the learned router with ranking based on motion magnitude. Learned routing improves mean PSNR by 1.44 dB and reduces LPIPS from 0.1018 to 0.0914, with 60.47 versus 62.50 FPS. Supplementary Sec.~D.4 also evaluates Residual-only and Combined heuristic, both of which yield lower image quality. Thus, learned update priorities preserve quality more effectively than these heuristics, at a small throughput cost. Across all twelve pairs, the router's mean PSNR is within 0.09 dB of the teacher's (Supplementary Tbl.~S3).

\textbf{Motion Prediction.}
We replace temporal extrapolation with hold-last prediction and train a router for this variant. Temporal extrapolation improves mean PSNR by 0.21 dB at nearly unchanged throughput, with 60.47 versus 60.84 FPS. This gain supports predicting unselected states from recent motion, consistent with the local regularity in \Sect{sec:method_motivation}.

\textbf{Scene-specific Budgets.}
Compared with a uniform 20\% update budget, calibrated scene budgets improve mean PSNR by 1.22 dB while reducing throughput from 67.57 to 60.47 FPS, a 10.5\% decrease. A common update fraction can allocate unnecessary work to some scenes while leaving too few corrections for others. These results support calibrating the budget according to each scene's prediction risk, as described in \Sect{sec:method_training}.

\textbf{History Refresh and Precomputation.}
On N3DV, enabling history refresh without precomputation raises mean PSNR from 25.50 to 25.71 dB and reduces LPIPS from 0.1803 to 0.1767, while average FPS decreases from 46.93 to 40.95. Supplementary Sec.~D.4 further analyzes how adaptive refresh protects difficult segments compared with fixed-interval refresh. Adding precomputation under the same refresh schedule maintains similar quality and raises FPS P5 from 17.68 to 26.20, a 48.2\% improvement. Average throughput remains nearly unchanged, decreasing by only 0.3\%, from 40.95 to 40.81 FPS. History refresh helps preserve quality, while precomputation improves slow-frame performance at similar average throughput.
\subsection{Sensitivity Study}
\label{sec:sensitivity}
\vspace{-5pt}

All sensitivity tests use E-D3DGS and average the nine N3DV scenes equally.

\begin{figure}[!tbp]
\centering
\input{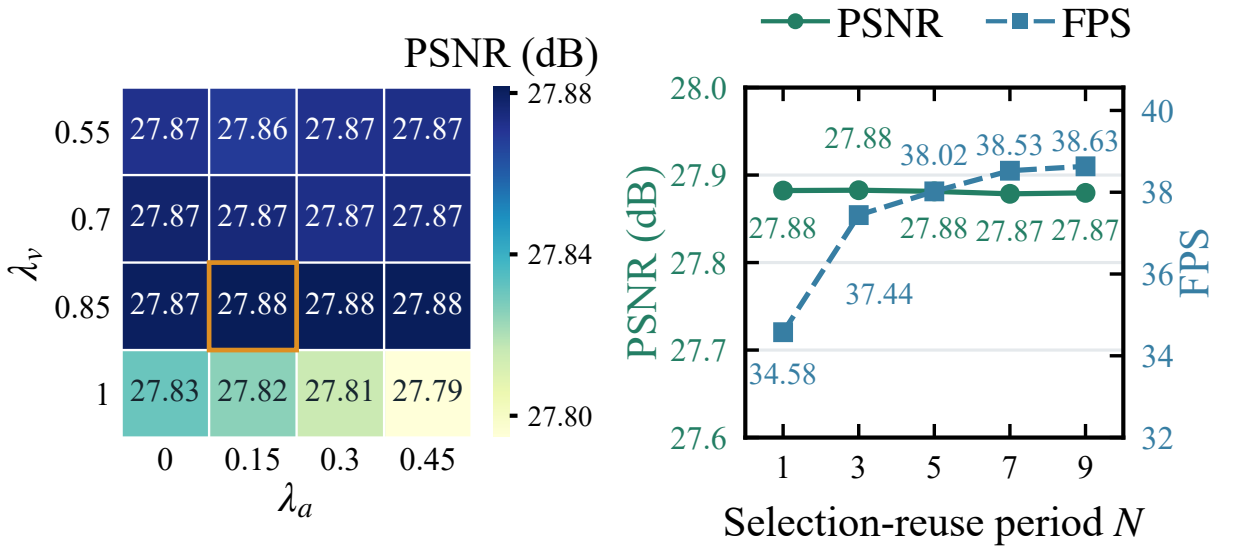}
\caption{Sensitivity of E-D3DGS on N3DV. (a) Mean PSNR across prediction coefficients, with the orange outline marking the default. (b) PSNR (left axis) and AGX Orin FPS (right axis) versus the selection-reuse period \(N\). PSNR labels are truncated to two decimal places.}
\label{fig:sensitivity}
\end{figure}

\needspace{4\baselineskip}
\textbf{Motion Prediction.}
\Fig{fig:sensitivity}(a) shows PSNR values of 27.86--27.88 dB across the tested damped settings. The default coefficients \((\lambda_v,\lambda_a)=(0.85,0.15)\) achieve 27.88 dB, compared with 27.83 dB for undamped linear prediction \((1,0)\). With \(\lambda_v=1\), increasing \(\lambda_a\) to 0.45 lowers PSNR to 27.79 dB. These results favor damping and show limited sensitivity within the tested damped range.

\textbf{Selection Reuse.}
Increasing \(N\) from 1 to 5 raises throughput from 34.58 to 38.02 FPS, a 10.0\% gain, while PSNR remains at 27.88 dB at the displayed precision (\Fig{fig:sensitivity}(b)). At \(N=7\) and 9, FPS improves by only 1.3\%--1.6\% over \(N=5\), with PSNR of 27.87 dB. The default period thus captures most of the measured speed benefit while limiting the interval between reselections.

\begin{table}[!tbp]
\centering
\caption{Effect of subsequent refresh length with initialization fixed to three frames, evaluated with E-D3DGS and averaged over nine N3DV scenes. FPS is measured on AGX Orin.}
\label{tab:refresh_length}
\small
\setlength{\tabcolsep}{2pt}
\begin{tabular*}{\columnwidth}{@{\extracolsep{\fill}}ccccc@{}}
\toprule
Refresh frames & PSNR (dB)$\uparrow$ & SSIM$\uparrow$ & LPIPS$\downarrow$ & FPS$\uparrow$ \\
\midrule
1 & 27.50 & 0.922 & 0.1184 & 41.83 \\
3 & 27.88 & 0.927 & 0.1144 & 38.36 \\
\bottomrule
\end{tabular*}
\end{table}

\needspace{4\baselineskip}
\textbf{Refresh Length.}
\Tbl{tab:refresh_length} compares one and three consecutive full-deformation frames per refresh after three-frame initialization. Using three frames raises PSNR from 27.50 to 27.88 dB and SSIM from 0.922 to 0.927, while reducing LPIPS from 0.1184 to 0.1144. FPS decreases from 41.83 to 38.36. This supports rebuilding all three states used by the history-based predictor in \Sect{sec:method_sparse}.

\section{Conclusion}
\label{sec:disc}

We presented \proj{}, a post-training framework that combines history-based prediction with learned update selection to accelerate dynamic Gaussian rendering.
A calibrated scene budget determines how many Gaussians receive exact updates, and the router adapts their selection over time.
History refreshes control accumulated prediction error, while precomputation improves slow-frame performance with little effect on average throughput.
Across four backbones and three datasets, \proj{} achieves $1.75$--$6.74\times$ speedups on Jetson AGX Orin, with dataset-average PSNR decreases of at most $0.16$~dB.
These results show that adapting deformation computation to local predictability can improve efficiency while preserving the pretrained representation.

The current implementation assumes sequential rendering at uniform time steps.
Future work will address temporal seeking and variable playback rates and reduce scene-specific preparation costs.

{\small
\bibliographystyle{ieeenat_fullname}
\bibliography{main}}

\begin{thebibliography}{54}
\providecommand{\natexlab}[1]{#1}
\providecommand{\url}[1]{\texttt{#1}}
\expandafter\ifx\csname urlstyle\endcsname\relax
  \providecommand{\doi}[1]{doi: #1}\else
  \providecommand{\doi}{doi: \begingroup \urlstyle{rm}\Url}\fi

\bibitem[Bae et~al.(2024)Bae, Kim, Yun, Lee, Bang, and Uh]{bae2024ed3dgs}
Jeongmin Bae, Seoha Kim, Youngsik Yun, Hahyun Lee, Gun Bang, and Youngjung Uh.
\newblock {Per-Gaussian Embedding-Based Deformation for Deformable 3D Gaussian
  Splatting}.
\newblock In \emph{ECCV}, pages 321--335, 2024.

\bibitem[Bewley et~al.(2016)Bewley, Ge, Ott, Ramos, and
  Upcroft]{bewley2016sort}
Alex Bewley, Zongyuan Ge, Lionel Ott, Fabio Ramos, and Ben Upcroft.
\newblock {Simple Online and Realtime Tracking}.
\newblock In \emph{{ICIP}}, pages 3464--3468, 2016.

\bibitem[Cao and Johnson(2023)]{cao2023hexplane}
Ang Cao and Justin Johnson.
\newblock {HexPlane: A Fast Representation for Dynamic Scenes}.
\newblock In \emph{CVPR}, pages 130--141, 2023.

\bibitem[Chen et~al.(2025)Chen, Hu, Wu, Zhu, Li, and Sun]{chen2025dash}
Jie Chen, Zhangchi Hu, Peixi Wu, Huyue Zhu, Hebei Li, and Xiaoyan Sun.
\newblock {DASH: 4D Hash Encoding with Self-Supervised Decomposition for
  Real-Time Dynamic Scene Rendering}.
\newblock In \emph{{ICCV}}, pages 26349--26359, 2025.

\bibitem[Chu et~al.(2020)Chu, Xie, Mayer, Leal-Taix{\'e}, and
  Thuerey]{chu2020tecogan}
Mengyu Chu, You Xie, Jonas Mayer, Laura Leal-Taix{\'e}, and Nils Thuerey.
\newblock {Learning Temporal Coherence via Self-Supervision for GAN-based Video
  Generation}.
\newblock \emph{ACM Transactions on Graphics}, 39\penalty0 (4), 2020.
\newblock Article 75.

\bibitem[Das et~al.(2024)Das, Wewer, Yunus, Ilg, and Lenssen]{das2024npg}
Devikalyan Das, Christopher Wewer, Raza Yunus, Eddy Ilg, and Jan~Eric Lenssen.
\newblock {Neural Parametric Gaussians for Monocular Non-Rigid Object
  Reconstruction}.
\newblock In \emph{CVPR}, pages 10715--10725, 2024.

\bibitem[Duan et~al.(2024)Duan, Wei, Dai, He, Chen, and Chen]{duan2024rotor}
Yuanxing Duan, Fangyin Wei, Qiyu Dai, Yuhang He, Wenzheng Chen, and Baoquan
  Chen.
\newblock {4D-Rotor Gaussian Splatting: Towards Efficient Novel View Synthesis
  for Dynamic Scenes}.
\newblock In \emph{ACM SIGGRAPH Conference Papers}, pages 1--11, 2024.

\bibitem[Dutson et~al.(2022)Dutson, Li, and Gupta]{dutson2022event}
Matthew Dutson, Yin Li, and Mohit Gupta.
\newblock {Event Neural Networks}.
\newblock In \emph{ECCV}, pages 276--293, 2022.

\bibitem[Dutson et~al.(2023)Dutson, Li, and Gupta]{dutson2023eventful}
Matthew Dutson, Yin Li, and Mohit Gupta.
\newblock {Eventful Transformers: Leveraging Temporal Redundancy in Vision
  Transformers}.
\newblock In \emph{{ICCV}}, pages 16911--16923, 2023.

\bibitem[Fan et~al.(2024)Fan, Wang, Wen, Zhu, Xu, and
  Wang]{fan2024lightgaussian}
Zhiwen Fan, Kevin Wang, Kairun Wen, Zehao Zhu, Dejia Xu, and Zhangyang Wang.
\newblock {LightGaussian: Unbounded 3D Gaussian Compression with 15x Reduction
  and 200+ FPS}.
\newblock In \emph{Advances in Neural Information Processing Systems}, pages
  140138--140158, 2024.

\bibitem[Fang and Wang(2024)]{fang2024mini}
Guangchi Fang and Bing Wang.
\newblock {Mini-Splatting: Representing Scenes with a Constrained Number of
  Gaussians}.
\newblock In \emph{ECCV}, pages 165--181, 2024.

\bibitem[Fridovich-Keil et~al.(2023)Fridovich-Keil, Meanti, Warburg, Recht, and
  Kanazawa]{fridovich2023kplanes}
Sara Fridovich-Keil, Giacomo Meanti, Frederik~Rahb{\ae}k Warburg, Benjamin
  Recht, and Angjoo Kanazawa.
\newblock {K-Planes: Explicit Radiance Fields in Space, Time, and Appearance}.
\newblock In \emph{CVPR}, pages 12479--12488, 2023.

\bibitem[Gao et~al.(2024)Gao, Meng, Wen, Chen, and Zhang]{gao2024hicom}
Qiankun Gao, Jiarui Meng, Chengxiang Wen, Jie Chen, and Jian Zhang.
\newblock {HiCoM: Hierarchical Coherent Motion for Dynamic Streamable Scenes
  with 3D Gaussian Splatting}.
\newblock In \emph{Advances in Neural Information Processing Systems}, pages
  80609--80633, 2024.

\bibitem[Girish et~al.(2024{\natexlab{a}})Girish, Gupta, and
  Shrivastava]{girish2024eagles}
Sharath Girish, Kamal Gupta, and Abhinav Shrivastava.
\newblock {EAGLES: Efficient Accelerated 3D Gaussians with Lightweight
  EncodingS}.
\newblock In \emph{ECCV}, pages 54--71, 2024{\natexlab{a}}.

\bibitem[Girish et~al.(2024{\natexlab{b}})Girish, Li, Mazumdar, Shrivastava,
  Luebke, and De~Mello]{girish2024queen}
Sharath Girish, Tianye Li, Amrita Mazumdar, Abhinav Shrivastava, David Luebke,
  and Shalini De~Mello.
\newblock {QUEEN: QUantized Efficient ENcoding of Dynamic Gaussians for
  Streaming Free-viewpoint Videos}.
\newblock In \emph{Advances in Neural Information Processing Systems}, pages
  43435--43467, 2024{\natexlab{b}}.

\bibitem[Hu et~al.(2025)Hu, Zheng, Zhong, Fu, Song, Zhang, Zhai, and
  Wang]{hu2025gc}
Qiang Hu, Zihan Zheng, Houqiang Zhong, Sihua Fu, Li Song, Xiaoyun Zhang,
  Guangtao Zhai, and Yanfeng Wang.
\newblock {4DGC: Rate-Aware 4D Gaussian Compression for Efficient Streamable
  Free-Viewpoint Video}.
\newblock In \emph{CVPR}, pages 875--885, 2025.

\bibitem[Huang et~al.(2024)Huang, Sun, Yang, Lyu, Cao, and Qi]{huang2024scgs}
Yi-Hua Huang, Yang-Tian Sun, Ziyi Yang, Xiaoyang Lyu, Yan-Pei Cao, and Xiaojuan
  Qi.
\newblock {SC-GS: Sparse-Controlled Gaussian Splatting for Editable Dynamic
  Scenes}.
\newblock In \emph{{CVPR}}, pages 4220--4230, 2024.

\bibitem[Katsumata et~al.(2024)Katsumata, Vo, and
  Nakayama]{katsumata2024compact}
Kai Katsumata, Duc~Minh Vo, and Hideki Nakayama.
\newblock {A Compact Dynamic 3D Gaussian Representation for Real-Time Dynamic
  View Synthesis}.
\newblock In \emph{ECCV}, pages 394--412, 2024.

\bibitem[Kerbl et~al.(2023)Kerbl, Kopanas, Leimk{\"u}hler, and
  Drettakis]{kerbl2023gs}
Bernhard Kerbl, Georgios Kopanas, Thomas Leimk{\"u}hler, and George Drettakis.
\newblock {3D Gaussian Splatting for Real-Time Radiance Field Rendering}.
\newblock \emph{ACM Transactions on Graphics}, 42\penalty0 (4):\penalty0 1--14,
  2023.

\bibitem[Kratimenos et~al.(2024)Kratimenos, Lei, and
  Daniilidis]{kratimenos2024dynmf}
Agelos Kratimenos, Jiahui Lei, and Kostas Daniilidis.
\newblock {DynMF: Neural Motion Factorization for Real-Time Dynamic View
  Synthesis with 3D Gaussian Splatting}.
\newblock In \emph{ECCV}, pages 252--269, 2024.

\bibitem[Kwak et~al.(2025)Kwak, Kim, Jeong, Cheong, Oh, and Kim]{kwak2025modec}
Sangwoon Kwak, Joonsoo Kim, Jun~Young Jeong, Won-Sik Cheong, Jihyong Oh, and
  Munchurl Kim.
\newblock {MoDec-GS: Global-to-Local Motion Decomposition and Temporal Interval
  Adjustment for Compact Dynamic 3D Gaussian Splatting}.
\newblock In \emph{{CVPR}}, pages 11338--11348, 2025.

\bibitem[Lee et~al.(2024)Lee, Won, Jung, Bae, and Jeon]{lee2024explicit}
Junoh Lee, Changyeon Won, Hyunjun Jung, Inhwan Bae, and Hae-Gon Jeon.
\newblock {Fully Explicit Dynamic Gaussian Splatting}.
\newblock In \emph{{Advances in Neural Information Processing Systems}}, pages
  5384--5409, 2024.

\bibitem[Li et~al.(2025)Li, Li, Gao, Batuer, Yu, and Liao]{li2025gifstream}
Hao Li, Sicheng Li, Xiang Gao, Abudouaihati Batuer, Lu Yu, and Yiyi Liao.
\newblock {GIFStream: 4D Gaussian-based Immersive Video with Feature Stream}.
\newblock In \emph{CVPR}, pages 21761--21770, 2025.

\bibitem[Li et~al.(2022)Li, Slavcheva, Zollh\"ofer, Green, Lassner, Kim,
  Schmidt, Lovegrove, Goesele, Newcombe, and Lv]{li2022neuralvideo}
Tianye Li, Mira Slavcheva, Michael Zollh\"ofer, Simon Green, Christoph Lassner,
  Changil Kim, Tanner Schmidt, Steven Lovegrove, Michael Goesele, Richard
  Newcombe, and Zhaoyang Lv.
\newblock {Neural 3D Video Synthesis From Multi-View Video}.
\newblock In \emph{{CVPR}}, pages 5521--5531, 2022.

\bibitem[Li et~al.(2024)Li, Chen, Li, and Xu]{li2024spacetime}
Zhan Li, Zhang Chen, Zhong Li, and Yi Xu.
\newblock {Spacetime Gaussian Feature Splatting for Real-Time Dynamic View
  Synthesis}.
\newblock In \emph{{CVPR}}, pages 8508--8520, 2024.

\bibitem[Lin et~al.(2024)Lin, Dai, Zhu, and Yao]{lin2024flow}
Youtian Lin, Zuozhuo Dai, Siyu Zhu, and Yao Yao.
\newblock {Gaussian-Flow: 4D Reconstruction with Dynamic 3D Gaussian Particle}.
\newblock In \emph{CVPR}, pages 21136--21145, 2024.

\bibitem[Liu et~al.(2017)Liu, Yeh, Tang, Liu, and Agarwala]{liu2017voxelflow}
Ziwei Liu, Raymond~A. Yeh, Xiaoou Tang, Yiming Liu, and Aseem Agarwala.
\newblock {Video Frame Synthesis Using Deep Voxel Flow}.
\newblock In \emph{{ICCV}}, pages 4463--4471, 2017.

\bibitem[Lu et~al.(2024)Lu, Guo, Hui, Chen, Yang, Tang, Zhu, and
  Dai]{lu2024gags}
Zhicheng Lu, Xiang Guo, Le Hui, Tianrui Chen, Min Yang, Xiao Tang, Feng Zhu,
  and Yuchao Dai.
\newblock {3D Geometry-Aware Deformable Gaussian Splatting for Dynamic View
  Synthesis}.
\newblock In \emph{{CVPR}}, pages 8900--8910, 2024.

\bibitem[Luiten et~al.(2024)Luiten, Kopanas, Leibe, and
  Ramanan]{luiten2024dynamic}
Jonathon Luiten, Georgios Kopanas, Bastian Leibe, and Deva Ramanan.
\newblock {Dynamic 3D Gaussians: Tracking by Persistent Dynamic View
  Synthesis}.
\newblock In \emph{{3DV}}, 2024.

\bibitem[Parger et~al.(2022)Parger, Tang, Twigg, Keskin, Wang, and
  Steinberger]{parger2022deltacnn}
Mathias Parger, Chengcheng Tang, Christopher~D. Twigg, Cem Keskin, Robert Wang,
  and Markus Steinberger.
\newblock {DeltaCNN: End-to-End CNN Inference of Sparse Frame Differences in
  Videos}.
\newblock In \emph{CVPR}, pages 12497--12506, 2022.

\bibitem[Parger et~al.(2023)Parger, Tang, Neff, Twigg, Keskin, Wang, and
  Steinberger]{parger2023motiondelta}
Mathias Parger, Chengcheng Tang, Thomas Neff, Christopher~D. Twigg, Cem Keskin,
  Robert Wang, and Markus Steinberger.
\newblock {MotionDeltaCNN: Sparse CNN Inference of Frame Differences in Moving
  Camera Videos with Spherical Buffers and Padded Convolutions}.
\newblock In \emph{ICCV}, pages 17292--17301, 2023.

\bibitem[Park et~al.(2025)Park, Bui, Bello, Moon, Oh, and Kim]{park2025spline}
Jongmin Park, Minh-Quan~Viet Bui, Juan Luis~Gonzalez Bello, Jaeho Moon, Jihyong
  Oh, and Munchurl Kim.
\newblock {SplineGS: Robust Motion-Adaptive Spline for Real-Time Dynamic 3D
  Gaussians from Monocular Video}.
\newblock In \emph{CVPR}, pages 26866--26875, 2025.

\bibitem[Park et~al.(2021{\natexlab{a}})Park, Sinha, Barron, Bouaziz, Goldman,
  Seitz, and Martin-Brualla]{park2021nerfies}
Keunhong Park, Utkarsh Sinha, Jonathan~T. Barron, Sofien Bouaziz, Dan~B
  Goldman, Steven~M. Seitz, and Ricardo Martin-Brualla.
\newblock {Nerfies: Deformable Neural Radiance Fields}.
\newblock In \emph{ICCV}, pages 5865--5874, 2021{\natexlab{a}}.

\bibitem[Park et~al.(2021{\natexlab{b}})Park, Sinha, Hedman, Barron, Bouaziz,
  Goldman, Martin-Brualla, and Seitz]{park2021hypernerf}
Keunhong Park, Utkarsh Sinha, Peter Hedman, Jonathan~T. Barron, Sofien Bouaziz,
  Dan~B Goldman, Ricardo Martin-Brualla, and Steven~M. Seitz.
\newblock {HyperNeRF: A Higher-Dimensional Representation for Topologically
  Varying Neural Radiance Fields}.
\newblock \emph{ACM Transactions on Graphics}, 40\penalty0 (6),
  2021{\natexlab{b}}.

\bibitem[Pumarola et~al.(2021)Pumarola, Corona, Pons-Moll, and
  Moreno-Noguer]{pumarola2021dnerf}
Albert Pumarola, Enric Corona, Gerard Pons-Moll, and Francesc Moreno-Noguer.
\newblock {D-NeRF: Neural Radiance Fields for Dynamic Scenes}.
\newblock In \emph{CVPR}, pages 10318--10327, 2021.

\bibitem[Sabater et~al.(2017)Sabater, Boisson, Vandame, Kerbiriou, Babon, Hog,
  Gendrot, Langlois, Bureller, Schubert, and Allie]{sabater2017dataset}
Neus Sabater, Guillaume Boisson, Benoit Vandame, Paul Kerbiriou, Frederic
  Babon, Matthieu Hog, Remy Gendrot, Tristan Langlois, Olivier Bureller, Arno
  Schubert, and Valerie Allie.
\newblock {Dataset and Pipeline for Multi-View Light-Field Video}.
\newblock In \emph{{CVPR Workshops}}, pages 1743--1753, 2017.

\bibitem[Shaw et~al.(2024)Shaw, Nazarczuk, Song, Moreau, Catley-Chandar, Dhamo,
  and P{\'e}rez-Pellitero]{shaw2024swings}
Richard Shaw, Michal Nazarczuk, Jifei Song, Arthur Moreau, Sibi Catley-Chandar,
  Helisa Dhamo, and Eduardo P{\'e}rez-Pellitero.
\newblock {SWinGS: Sliding Windows for Dynamic 3D Gaussian Splatting}.
\newblock In \emph{ECCV}, pages 37--54, 2024.

\bibitem[Sun et~al.(2024)Sun, Jiao, Li, Zhang, Zhao, and Xing]{sun2024stream}
Jiakai Sun, Han Jiao, Guangyuan Li, Zhanjie Zhang, Lei Zhao, and Wei Xing.
\newblock {3DGStream: On-the-Fly Training of 3D Gaussians for Efficient
  Streaming of Photo-Realistic Free-Viewpoint Videos}.
\newblock In \emph{{CVPR}}, pages 20675--20685, 2024.

\bibitem[Tu et~al.(2026)Tu, Ying, Hanson, Lee, Goldstein, and
  Zwicker]{tu2026speede}
Allen Tu, Haiyang Ying, Alex Hanson, Yonghan Lee, Tom Goldstein, and Matthias
  Zwicker.
\newblock {SpeeDe3DGS: Speedy Deformable 3D Gaussian Splatting with Temporal
  Pruning and Motion Grouping}.
\newblock In \emph{{CVPR}}, pages 26083--26093, 2026.

\bibitem[Wang et~al.(2024)Wang, Zhang, Wang, Yao, Xie, Yu, Wu, and
  Xu]{wang2024v3}
Penghao Wang, Zhirui Zhang, Liao Wang, Kaixin Yao, Siyuan Xie, Jingyi Yu, Minye
  Wu, and Lan Xu.
\newblock {V$^3$: Viewing Volumetric Videos on Mobiles via Streamable 2D
  Dynamic Gaussians}.
\newblock \emph{ACM Transactions on Graphics}, 43\penalty0 (6):\penalty0 1--13,
  2024.

\bibitem[Wang et~al.(2025)Wang, Yang, Xu, Sun, Zhang, Chen, Bao, Peng, and
  Zhou]{wang2025freetime}
Yifan Wang, Peishan Yang, Zhen Xu, Jiaming Sun, Zhanhua Zhang, Yong Chen, Hujun
  Bao, Sida Peng, and Xiaowei Zhou.
\newblock {FreeTimeGS: Free Gaussian Primitives at Anytime Anywhere for Dynamic
  Scene Reconstruction}.
\newblock In \emph{CVPR}, pages 21750--21760, 2025.

\bibitem[Wang et~al.(2004)Wang, Bovik, Sheikh, and Simoncelli]{wang2004ssim}
Zhou Wang, Alan~C. Bovik, Hamid~R. Sheikh, and Eero~P. Simoncelli.
\newblock {Image Quality Assessment: From Error Visibility to Structural
  Similarity}.
\newblock \emph{IEEE Transactions on Image Processing}, 13\penalty0
  (4):\penalty0 600--612, 2004.

\bibitem[Wu et~al.(2024)Wu, Yi, Fang, Xie, Zhang, Wei, Liu, Tian, and
  Wang]{wu2024four}
Guanjun Wu, Taoran Yi, Jiemin Fang, Lingxi Xie, Xiaopeng Zhang, Wei Wei, Wenyu
  Liu, Qi Tian, and Xinggang Wang.
\newblock {4D Gaussian Splatting for Real-Time Dynamic Scene Rendering}.
\newblock In \emph{{CVPR}}, pages 20310--20320, 2024.

\bibitem[Xu et~al.(2026)Xu, Duan, Dai, Li, Chen, and Wang]{xu2026layered}
Hanjie Xu, Yuanxing Duan, Qiyu Dai, Ge Li, Baoquan Chen, and He Wang.
\newblock {Layered 4D-Rotor Gaussian Splatting: A Compressed Representation for
  Long Dynamic Scenes}.
\newblock In \emph{CVPR}, pages 18958--18967, 2026.

\bibitem[Xu et~al.(2024)Xu, Xu, Yu, Peng, Sun, Bao, and Zhou]{xu2024hierarchy}
Zhen Xu, Yinghao Xu, Zhiyuan Yu, Sida Peng, Jiaming Sun, Hujun Bao, and Xiaowei
  Zhou.
\newblock {Representing Long Volumetric Video with Temporal Gaussian
  Hierarchy}.
\newblock \emph{ACM Transactions on Graphics}, 43\penalty0 (6):\penalty0 1--18,
  2024.

\bibitem[Yan et~al.(2025)Yan, Peng, Wang, Tang, Yang, Liang, Wu, and
  Wang]{yan2025instant}
Jinbo Yan, Rui Peng, Zhiyan Wang, Luyang Tang, Jiayu Yang, Jie Liang, Jiahao
  Wu, and Ronggang Wang.
\newblock {Instant Gaussian Stream: Fast and Generalizable Streaming of Dynamic
  Scene Reconstruction via Gaussian Splatting}.
\newblock In \emph{CVPR}, pages 16520--16531, 2025.

\bibitem[Yan et~al.(2023)Yan, Li, and Lee]{yan2023nerfds}
Zhiwen Yan, Chen Li, and Gim~Hee Lee.
\newblock {NeRF-DS: Neural Radiance Fields for Dynamic Specular Objects}.
\newblock In \emph{{CVPR}}, pages 8285--8295, 2023.

\bibitem[Yang et~al.(2024{\natexlab{a}})Yang, Gao, Zhou, Jiao, Zhang, and
  Jin]{yang2024deformable}
Ziyi Yang, Xinyu Gao, Wen Zhou, Shaohui Jiao, Yuqing Zhang, and Xiaogang Jin.
\newblock {Deformable 3D Gaussians for High-Fidelity Monocular Dynamic Scene
  Reconstruction}.
\newblock In \emph{{CVPR}}, pages 20331--20341, 2024{\natexlab{a}}.

\bibitem[Yang et~al.(2024{\natexlab{b}})Yang, Yang, Pan, and
  Zhang]{yang2024native4d}
Zeyu Yang, Hongye Yang, Zijie Pan, and Li Zhang.
\newblock {Real-time Photorealistic Dynamic Scene Representation and Rendering
  with 4D Gaussian Splatting}.
\newblock In \emph{{ICLR}}, pages 9142--9159, 2024{\natexlab{b}}.

\bibitem[Ye et~al.(2025)Ye, Li, Kerr, Turkulainen, Yi, Pan, Seiskari, Ye, Hu,
  Tancik, and Kanazawa]{ye2025gsplat}
Vickie Ye, Ruilong Li, Justin Kerr, Matias Turkulainen, Brent Yi, Zhuoyang Pan,
  Otto Seiskari, Jianbo Ye, Jeffrey Hu, Matthew Tancik, and Angjoo Kanazawa.
\newblock {gsplat: An Open-Source Library for Gaussian Splatting}.
\newblock \emph{Journal of Machine Learning Research}, 26\penalty0
  (34):\penalty0 1--17, 2025.

\bibitem[Zhang et~al.(2018)Zhang, Isola, Efros, Shechtman, and
  Wang]{zhang2018lpips}
Richard Zhang, Phillip Isola, Alexei~A. Efros, Eli Shechtman, and Oliver Wang.
\newblock {The Unreasonable Effectiveness of Deep Features as a Perceptual
  Metric}.
\newblock In \emph{{CVPR}}, pages 586--595, 2018.

\bibitem[Zhang et~al.(2026)Zhang, Chen, Liu, Wu, and Chen]{zhang2026catgs}
Tingjia Zhang, Bo Chen, Shengzhong Liu, Fan Wu, and Guihai Chen.
\newblock {CaT-GS: Efficient 3DGS Rendering for Large-Scale Scenes with
  Inter-frame Caching and Tile Scheduling}.
\newblock In \emph{{CVPR}}, pages 37391--37400, 2026.

\bibitem[Zhang et~al.(2025)Zhang, Liu, Zhang, Ge, He, Xu, Wang, Lin, Yan, and
  Zhang]{zhang2025mega}
Xinjie Zhang, Zhening Liu, Yifan Zhang, Xingtong Ge, Dailan He, Tongda Xu, Yan
  Wang, Zehong Lin, Shuicheng Yan, and Jun Zhang.
\newblock {MEGA: Memory-Efficient 4D Gaussian Splatting for Dynamic Scenes}.
\newblock In \emph{ICCV}, pages 27828--27838, 2025.

\bibitem[Zhao et~al.(2024)Zhao, Li, Sun, Zeng, Shen, Ma, Zhang, Bao, and
  Cui]{zhao2024prediction}
Boming Zhao, Yuan Li, Ziyu Sun, Lin Zeng, Yujun Shen, Rui Ma, Yinda Zhang,
  Hujun Bao, and Zhaopeng Cui.
\newblock {GaussianPrediction: Dynamic 3D Gaussian Prediction for Motion
  Extrapolation and Free View Synthesis}.
\newblock In \emph{ACM SIGGRAPH Conference Papers}, pages 1--12, 2024.

\end{thebibliography}
\end{document}